\pdfoutput=1

\documentclass[11pt]{article}

\usepackage[]{eacl2023}

\usepackage{times}
\usepackage{latexsym}

\usepackage[T1]{fontenc}

\usepackage[utf8]{inputenc}

\usepackage{microtype}

\usepackage{booktabs}
\usepackage{graphicx}
\usepackage{amsmath,amssymb,amsfonts}
\usepackage[titlenumbered,ruled]{algorithm2e}
\usepackage{csquotes}
\usepackage{pifont}
\usepackage{bbm}
\usepackage[noabbrev]{cleveref}
\usepackage{caption}
\usepackage{subcaption}
\usepackage{tikz}

\newcommand{\alanno}{\textsc{alanno}}

\newcommand*\circled[1]{\tikz[baseline=(char.base)]{
            \node[shape=circle,draw,inner sep=1pt] (char) {#1};}}

\title{\alanno{}: An Active Learning Annotation System for Mortals}

\author{Josip Juki{\'{c}}$^{\spadesuit}$ \quad Fran Jeleni{\'{c}}$^{\spadesuit}$ \quad Miroslav Bi{\'{c}}ani{\'{c}} \quad Jan {\v{S}}najder\\
University of Zagreb, Faculty of Electrical Engineering and Computing\\
Text Analysis and Knowledge Engineering Lab\\
\tt \{name.surname\}@fer.hr
} 

\begin{document}
\maketitle
\def\thefootnote{$\spadesuit$}\footnotetext{Equal contribution}
\def\thefootnote{\arabic{footnote}}
\begin{abstract}

Supervised machine learning has become the cornerstone of today’s data-driven society, increasing the need for labeled data. However, the process of acquiring labels is often expensive and tedious. One possible remedy is to use \textit{active learning} (AL) -- a special family of machine learning algorithms designed to reduce labeling costs. Although AL has been successful in practice, a number of practical challenges hinder its effectiveness and are often overlooked in existing AL annotation tools. To address these challenges, we developed \alanno{}, an open-source annotation system for NLP tasks equipped with features to make AL effective in real-world annotation projects. \alanno{} facilitates annotation management in a multi-annotator setup and supports a variety of AL methods and underlying models, which are easily configurable and extensible.

\end{abstract}

\section{Introduction}

We are witnessing an ever-growing demand for data along with the rapid development of machine learning and deep learning algorithms. In particular, we need an abundance of labeled data to develop well-performing models, which is not easy to obtain. For many natural language processing (NLP) tasks, the labeling process, i.e., annotation, is often the most expensive and time-consuming part of developing machine learning models. The cognitive exertion of human annotators can affect their judgment, which further 
 affects label validity. Consequently, this manifests in poor \textit{agreement} -- a proxy for label reliability, which is a prerequisite for validity \cite{artstein-poesio-2008-inter, paun-et-al-2022-statistical}. Poor label reliability and validity negatively affect the machine learning algorithm, as it is only as good as the data it consumes.

Designed to alleviate labeling issues and reduce annotation cost, \emph{active learning} \cite[\textbf{AL};][]{settles-2009-active} is a special family of machine learning algorithms. In contrast to the standard random selection of instances for labeling, a typical AL method iteratively queries the most informative instances for the underlying model to achieve the best possible performance with the fewest possible labels. AL has been shown to reduce annotation effort across machine learning applications, e.g., \cite{beluch-et-al-2018-power, zhang-chen-2002-active}, especially in NLP, e.g., \cite{chen-et-al-2012-applying, settles-craven-2008-analysis, ein-dor-etal-2020-active}.

Despite the demonstrated successes of AL, many challenges are involved in deploying AL in real-world scenarios \cite{lowell-et-al-2018-practical, attenberg-provost-2011-inactive}. Unfortunately, these challenges are often overlooked in both research and practice. In particular, annotation tools that support AL rarely address the problems of unbiased evaluation of AL, imbalanced data, and stopping criteria for AL. The lack of concrete solutions for these problems hinders the effectiveness of AL. Aside from the practical challenges in AL, managing annotation campaigns is often very cumbersome, especially in multi-annotator setups (when multiple annotators are assigned to a single instance). Specifically, assigning instances to multiple annotators can be painstaking, particularly if one aims to achieve balanced combinations of annotators across instances. While there are many serviceable frameworks for simulating AL in idealized scenarios, e.g., \cite{danka-horvath-2018-modal, tang-etal-2019-alipy, schroder-etal-2021-small}, there are only a few tools for running real-world AL annotation campaigns with multiple annotators, none of them explicitly addressing the practical AL challenges.

To facilitate the creation of high-quality NLP datasets at reduced annotation costs, we developed \alanno{} (\textbf{A}ctive \textbf{L}earning \textbf{Anno}tation), an open-source annotation system with AL strategies for data sampling. 
\alanno{}'s is specifically designed to address the practical challenges of AL and facilitate the management of multi-annotator annotation projects.
In particular, \alanno{} guides toward more quality labels with a novel method for the balanced assignment of unlabeled instances to annotators in a multi-annotator setup. We support building gold labels by monitoring the inter-annotator agreement with task-specific metrics and agreement-aware weighted aggregation of labels. Equally important, \alanno{} incorporates many features to address the major challenges of using AL in practice. Namely, we support guided learning \cite{attenberg-provost-2010-label} for mitigating data imbalance, and we ensure trustworthy evaluation of the underlying model on an unbiased test set and a stopping criterion to maximize the effectiveness of AL. As an essential practical solution, we enable a project-specific stopping criterion with a novel performance forecasting method based on Bayesian regression. By estimating the performance of the underlying model with hypothetically enlarged labeled sets, we enable practitioners to determine on the spot whether further annotation will only have diminishing returns. Lastly, \alanno{} supports a wide range of state-of-the-art AL methods from the literature, allowing seamless inclusion of new models or methods. 

In summary, our main contribution is \alanno{}, an open-source AL annotation system for NLP tasks, which features (1) practical strategies for applying AL to real-world problems with a range of AL methods and (2) annotation management facilitation in a multi-annotator setup with quality control. \alanno{} enables non-experts in AL to reap its benefits by accounting for key practical issues in annotation management and AL. In two case studies, we demonstrate \alanno{}'s two key features -- balanced data assignment and AL performance forecasting. We also provide a short video\footnote{\url{https://www.youtube.com/watch?v=hPcHPM8ttvE}} demonstration and release the code\footnote{\url{https://github.com/josipjukic/alanno}} under the Apache 2.0 license. While \alanno{} has been born out of several years of experience with NLP annotations for various tasks and has evolved with each new project, it remains highly configurable, allowing easy customization and extension.
\section{System Overview}
\label{sec:overview}

We briefly describe the key aspects of \alanno{}, which include projects, data assignment, label management, and annotation.

\begin{figure}[t!]
\centering
\includegraphics[width=\linewidth]{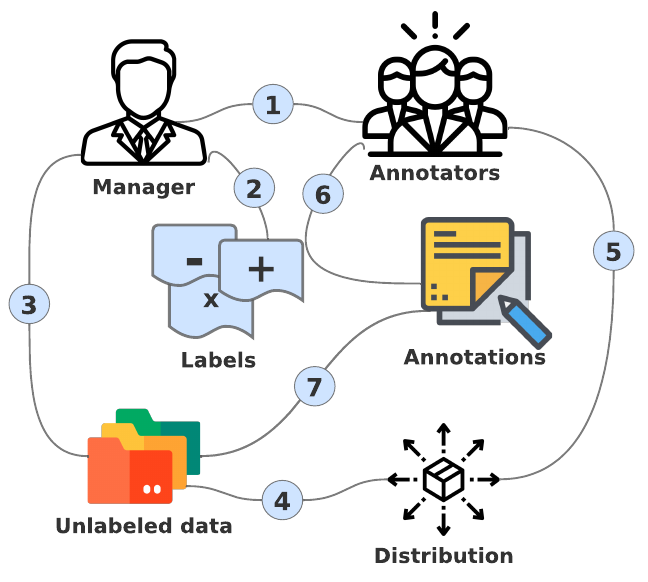}
\caption{Organization of an \alanno{} project. The lines between the icons indicate different lines of interaction, with the numbers denoting the temporal order. In brief, the project manager recruits annotators \circled{1}, creates labels \circled{2}, and imports the unlabeled data \circled{3}, which are then appointed to annotators via an assignment algorithm \circled{4}, \circled{5}. The annotators use the created labels \circled{6} to annotate the data \circled{7}.}
\label{fig:organization}
\end{figure}

\paragraph{Projects.}
In \alanno{}, the entire annotation process is encapsulated into a \emph{project}, which handles the interactions between different parts of an annotation project, as depicted in \Cref{fig:organization}. A typical NLP annotation project is  long-lasting and dynamic: annotators may be temporarily unavailable, new annotators may join an already-running project, and others may leave. \alanno{} supports the managing of such a workforce dynamic. To separate the concerns and responsibilities, \alanno{} defines two user roles: \textbf{project managers}, who are in charge of the annotation campaign, and \textbf{annotators}, whose task is to apply labels to the unlabeled data. At the moment, project managers can create three main types of projects: single- and multi-label classification, as well as sequence labeling tasks (e.g., named entity recognition).

\begin{figure}
\centering
\includegraphics[width=\linewidth]{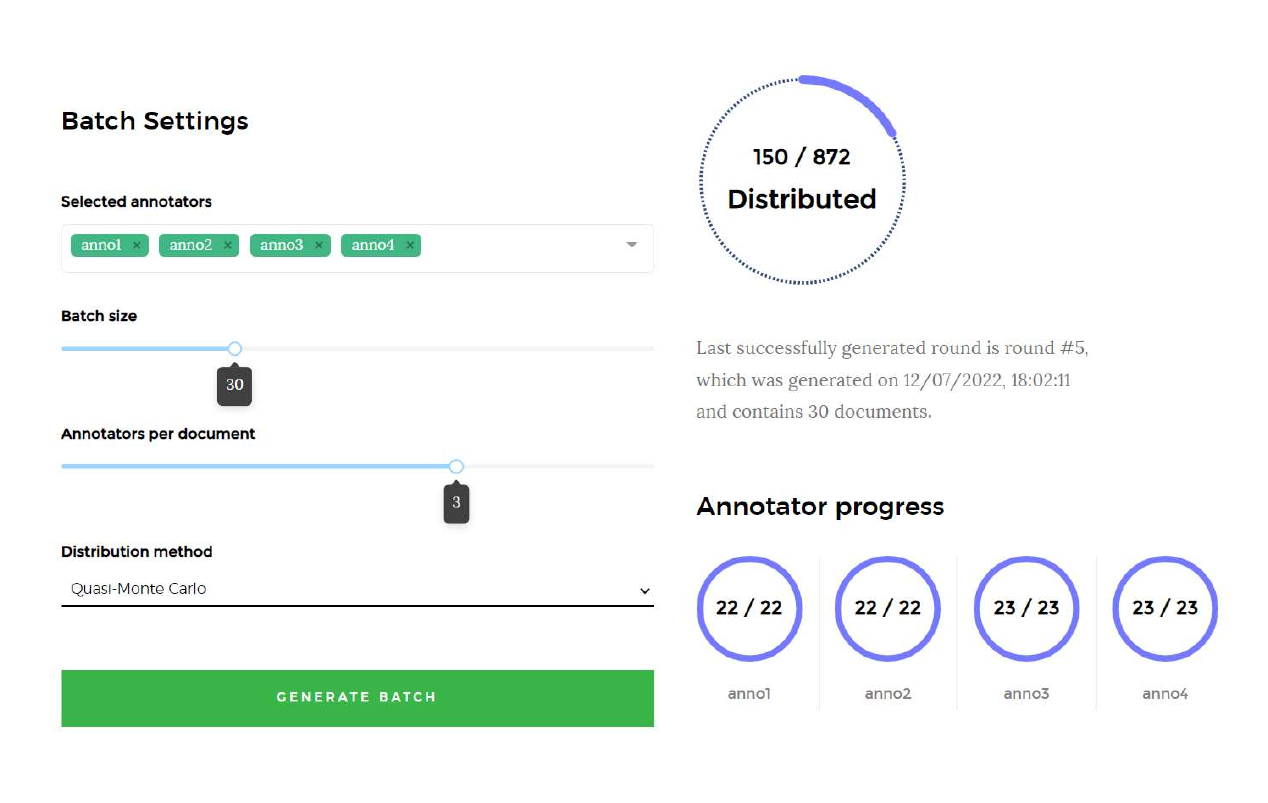}
\caption{Data assignment interface}
\label{fig:assignment}
\end{figure}

\paragraph{Data assignment.}
 Due to the dynamic nature of real-world annotation campaigns, it is convenient to separate the annotation process into smaller chunks. Moreover, annotation is an incremental process that often requires calibration in the initial phases. To meet these needs, the workload in \alanno{} is divided into rounds, where each round can be configured independently. The project manager can specify the number of unlabeled instances to be assigned, select annotators for the round, and the number of annotators per instance (\Cref{fig:assignment}).

\begin{figure}
\centering
\includegraphics[width=\linewidth]{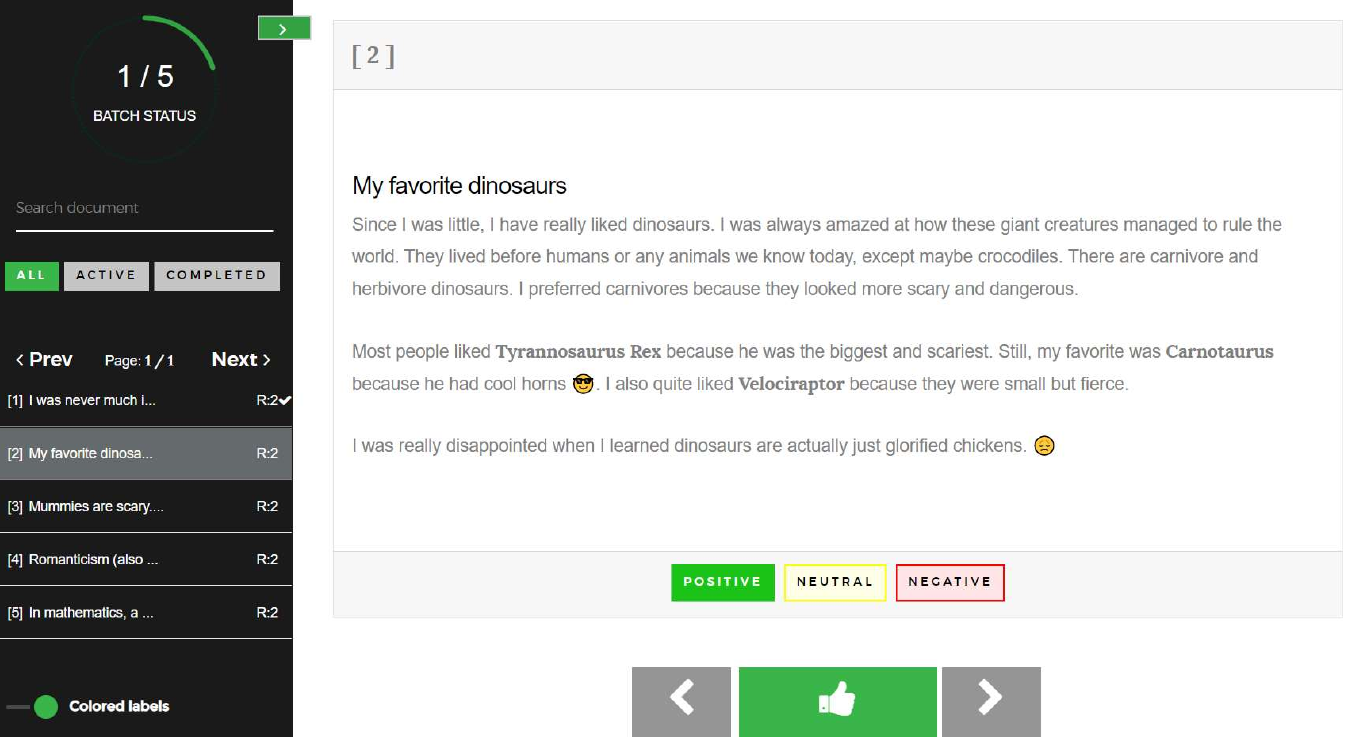}
\caption{Annotation interface}
\label{fig:annotation}
\end{figure}

\paragraph{Annotation.}
The annotation interface (\Cref{fig:annotation}) depends on the project type. For \emph{classification} tasks, we support a single-label setup, where only one of the labels can be applied, and a multi-label setup with the possibility of applying multiple labels. In addition, we cover a large variety of \emph{sequence labeling} tasks, where it is only necessary to define labels to fit the context of a specific use case. For example, one can define \emph{organization}, \emph{person name}, and \emph{location} as labels for named entity recognition. Annotators can then select spans of text that fall into one of the defined categories.  

\paragraph{Data management.}
The end product of an \alanno{} project is the annotated dataset, which consists of labels gathered from annotators. \alanno{} offers the user a choice between exporting an aggregated dataset, in which each instance appears precisely once with a single label obtained by aggregating the labels of different annotators, or a complete dataset, where each instance appears as many times as it has been annotated. The latter option is in line with recent recommendations to publish annotated datasets with the original labels rather than adjudicated labels \cite{kenyon-dean-etal-2020-deconstructing}, allowing for disagreement analysis, training of models that predict soft labels, e.g., \cite{pavlick2019inherent}, or application of statistical label aggregation techniques, e.g., \cite{qing-2014-empirical, hovy-yang-2021-importance, gordon2021disagreement}. Furthermore, to make it possible to follow up on earlier annotation projects, \alanno{} supports partially annotated datasets where the annotations are specified as user-label pairs.
\section{Features}
\label{sec:features}

Motivated by our experience in annotation projects and the practical challenges that emerge when deploying AL, we designed practical solutions that enable efficient labeling in real-world scenarios. We identified several key challenges, which, if not adequately addressed, may impair label quality and AL efficiency. Specifically, we focused on (1) label reliability, (2) unbiased evaluation of active learning models, (3) the stopping criterion for active learning, i.e., knowing when to terminate the active learning process, and (4) working with imbalanced data.

\subsection{Annotation management}

Annotation management in \alanno{} is centered around the first critical challenge -- label reliability. We support agreement-aware label aggregation and advanced data assignment to simultaneously promote label quality and make the management of annotation campaigns as seamless as possible.

\paragraph{Balanced assignment.}
Assigning unlabeled instances to annotators is an important aspect of the annotation process, in which combinations of annotators assigned to particular data points should be balanced to achieve more reliable labels. We have found that using uniform sampling based on pseudo-random numbers results in unbalanced combinations of annotators, with varying frequencies of different annotator tuples. To mitigate this and help improve label reliability, we developed a \emph{Quasi-Monte Carlo} assignment method based on quasi-random numbers. In particular, we used Sobol sequences \cite{burhenne-et-al-2011-sampling} to produce balanced combinations. In a scenario with $n$ annotators and $k$ annotators per data point, we draw $n$ dimensional vectors generated from the Sobol sequence, where each element is in the $[0, 1]$ interval. We round the values to the nearest integer ($0$ or $1$). If a particular vector has exactly $k$ elements with value $1$, we distribute the data point to the annotators at the corresponding indices of the vector. Otherwise, we discard the vector and draw a new one. The process is guaranteed to converge since all possible combinations are covered in the first $2^n$ vectors from the sequence. This procedure produces balanced combinations with uniform frequencies of annotator pairs, triplets, and up to $k$-tuples. We demonstrate its effects in a case study in \Cref{subsec:cs1}.

\paragraph{Monitoring agreement.}
In a multi-annotator setup, annotator agreement is a strong indicator of label quality. \alanno{} computes the inter-annotator agreement using metrics appropriate for the particular NLP task. Specifically, we use Cohen's $\kappa$ coefficient \cite{cohen-1960-coefficient} to evaluate pairwise agreement for binary and multi-class annotation. For the joint measure that considers all annotators simultaneously, we use Fleiss' $\kappa$ \cite{fleiss-1971-measuring}. On the other hand, for the multi-label setup, we use Krippendorff's $\alpha$ coefficient \cite{krippendorff-2018-content} paired with MASI distance \cite{passonneau-2006-measuring} for both pairwise and joint agreement.

\paragraph{Gold labels.}
Aggregating labels from multiple annotators is a critical component of creating high-quality datasets. In practice, different annotators often have different reliability levels due to differences in expertise. Such differences are exceptionally prominent with large groups of annotators. Therefore, \alanno{} generates \emph{gold labels} that consider an estimate of annotators' reliability. In particular, we aggregate the labels by assigning each annotator a weight proportional to how many times they assigned the majority label to a data point \cite{qing-2014-empirical}. For tasks with multiple labels, we chose to use the majority principle. The weighted aggregation leaves room for future improvement by incorporating systems such as Multi-Annotator Competence Estimation \cite[\textsc{mace};][]{hovy-etal-2013-learning}.

\subsection{AL acquisition models and functions}

\alanno{} supports AL as one of the key features. We incorporate practical solutions to mitigate the problems of deploying AL in real-world scenarios. We first describe what the system offers in terms of \emph{acquisition models}, i.e., the underlying models used for AL, and \emph{acquisition functions}, i.e., AL methods.

\alanno{} offers a rich palette of acquisition models for AL. We include various approaches to preprocessing tailored for a specific language for the NLP task at hand, including TF-IDF, customizable $n$-gram models, and word embeddings, using English as the default language. Besides English, we currently also support Croatian. \alanno{} provides many traditional models, including logistic regression, SVM, and random forest classifier. We also support deep models such as recurrent networks and Transformers \cite{vaswani-2017-attention}.

\alanno{} supports a wide range of active acquisition functions for both traditional and deep learning models. Starting from uncertainty sampling \cite{settles-2009-active}, a simple but powerful family of AL methods, \alanno{} covers the \emph{least confident}, \emph{margin}, and \emph{entropy} methods. All uncertainty-based methods are available for single- and multi-label problems. We have also incorporated acquisition functions that focus more on data diversity, such as the \emph{informative density} method, which leverages information about the instances in the input space and gives higher weights to instances in high-density parts of the input space. From the family of AL methods specialized for deep neural networks, \alanno{} provides the \emph{core-set} method \cite{sener-2017-active} and \textsc{badge} \cite{ash-etal-2019-deep}.

\subsection{AL challenges and solutions}

We describe the aforementioned practical challenges in AL (unbiased evaluation, stopping criterion, and class imbalance) and our solutions that aim to preserve AL effectiveness.

\paragraph{Unbiased evaluation.}
Before starting the annotation process, \alanno{} reserves a random sample of the imported data to be used later as a test set. In each round, managers can select how many test instances drawn from the reserved pool should be labeled out of the entire batch. In this way, one can adequately evaluate the model, as the reserved pool is not affected by the sampling bias \cite{prabhu2019sampling}. Since the acquisition functions often rely on the acquisition model's output, it is important to decouple evaluation and AL selection. A biased test set can lead to overestimating the model's performance, establishing a vicious cycle of uninformative queries in the early stages of acquisition. This often leads to redundant labels and, consequently, poorly performing models \cite{attenberg-provost-2011-inactive}.

\begin{figure}
\centering
\includegraphics[width=\linewidth]{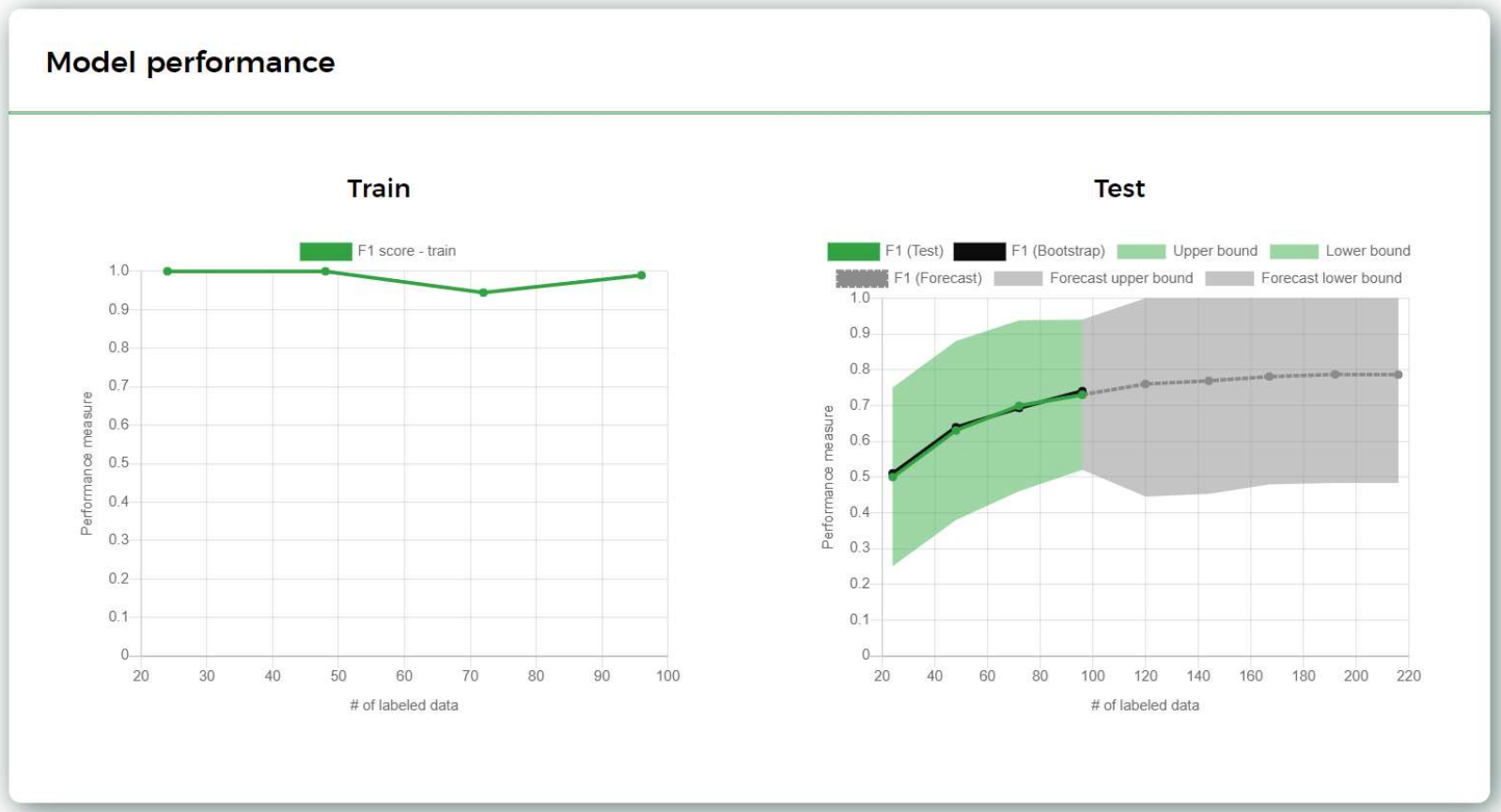}
\caption{AL evaluation and performance forecasting. After each round, \alanno{} re-trains the underlying AL model and plots the corresponding performance on the train set (the plot on the left-hand side). The performance is calculated on an unbiased test set reserved beforehand. We use $F_1$ score for the classification tasks, with confidence intervals approximated by bootstrapping. The plot on the right-hand side shows test performances on the data labeled so far (in green) and the forecast of performance increase with additional annotation effort (in gray).}
\label{fig:forecast}
\end{figure}

\paragraph{Stopping criterion.}
Although several stopping criteria for active learning have been proposed \cite{vlachos-2008-stopping, zhu-et-al-2010-confidence, laws-hinrich-2008-stopping, bloodgood-shanker-2014-method}, they are rarely employed in practice. We can save valuable resources by using a stopping criterion to identify when our model performs sufficiently well. Even more helpful would be the ability to forecast how much more data needs to be annotated to reach the desired model performance. To this end, we integrated a forecasting feature based on Bayesian regression, which we have implemented in Pyro \cite{bingham-2019-pyro}. The forecasting functionality is a practical solution for the stopping criterion, allowing annotation managers to gauge the trade-off between the expected boost in performance versus the additional annotation effort. \Cref{fig:forecast} shows an example of evaluating an active learning model and performance forecasting.

\paragraph{Class balancing through guided learning.}
AL often struggles with imbalanced data. Moreover, class balance is crucial for active learning strategies, especially in the early phases of the annotation process, as the model may have difficulty learning classes with low frequency. To address this, \alanno{} supports \emph{guided learning}, also known as \emph{active search} \cite{attenberg-provost-2010-label}. The main idea of guided learning in NLP is to use keywords to search for data points in the minority class. This way, users can annotate the retrieved data to make the class frequency distribution more uniform. We use BM25 \cite{robertson-2009-probabilistic} as the retrieval algorithm for guided learning.
\section{Case Studies}
\label{sec:cs}

In the following case studies, we highlight the two essential features of \alanno{}, namely balanced data assignment and AL performance forecasting.

\paragraph{Case study 1: Balanced assignment.}
\label{subsec:cs1}
To compare our Quasi-Monte Carlo assignment method with uniform annotators combinations, we ran simulations of distributing unlabeled instances to annotators. As \Cref{tab:var} shows, our method achieves more balanced combinations compared to the standard uniform sampling, ranging from pairs and up to $k$-tuples, where $k$ is the number of annotators per data point.

\begin{table}[tb!]
\small
\centering
\begin{tabular}{lrrrr}
\toprule
& $k=2$ & $k=3$ & $k=4$ & $k=5$ \\
\midrule
\textsc{uniform} & $132.66$ & $52.14$ & $18.09$ & $3.38$ \\
\textsc{qmc} & $0.75$ & $0.42$ & $0.15$ & $.03$ \\
\bottomrule
\end{tabular}
\caption{The average variance of $k$-tuple frequencies. \textsc{uniform} denotes the standard uniform sampling of annotators, while \textsc{qmc} stands for our Quasi-Monte Carlo assignment method. We simulated the assignment of 1,000 unlabeled instances with ten annotators in total and five annotators per instance. We report the average variance of frequencies across 1,000 runs.}
\label{tab:var}
\end{table}

\begin{figure*}[t!]
\small
\centering
\begin{subfigure}{.49\linewidth}
  \centering
  \includegraphics[width=\linewidth]{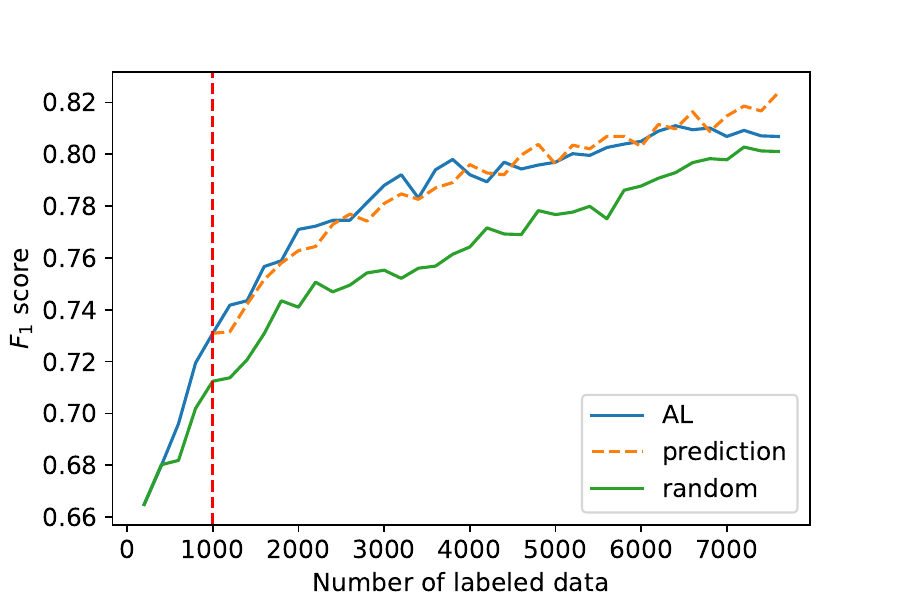}
\end{subfigure}
\begin{subfigure}{.49\linewidth}
  \centering
  \includegraphics[width=\linewidth]{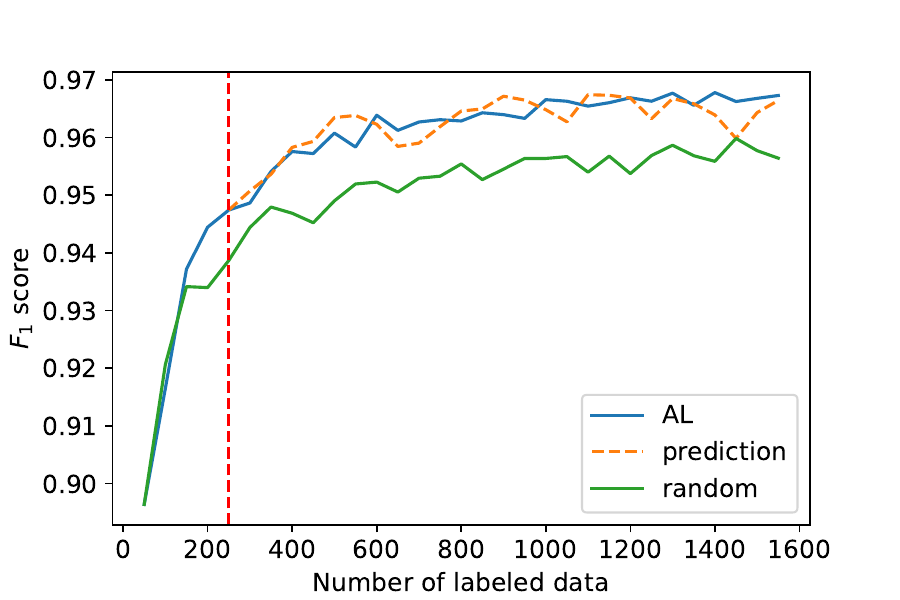}
\end{subfigure}
\caption{AL performance forecasting. The subfigures show the predictions of $F_1$ score gain with a hypothetical increase in the number of labeled data. The green line pertains to the random selection baseline, the blue one is the corresponding AL selection, and the dashed orange line is the forecast of $F_1$ score for AL selection. The dashed red line represents the boundary between the steps the forecast model was trained on (left of the line) and the steps whose values the forecast model was predicting (right of the line). The left subfigure shows the results for \textsc{sst} with logistic regression using uncertainty sampling as the AL method, while the right subfigure corresponds to \textsc{subj} with \textsc{bert} using \textsc{badge} for AL sampling.}
\label{fig:case-study}
\end{figure*}

\paragraph{Case study 2: AL performance forecasting.}
\label{subsec:cs2}
To demonstrate the forecasting feature in \alanno{}, we conducted a case study on the Stanford Sentiment Treebank \cite[\textbf{\textsc{sst};}][]{socher-etal-2013-parsing} and subjectivity \cite[\textbf{\textsc{subj};}][]{pang-lee-2004-sentimental} datasets. We used a simple logistic regression model with TF-IDF vectors and \emph{least confident} sampling method for \textsc{sst} and \textsc{bert} with \textsc{badge} sampling method for \textsc{subj}. We then compared random sampling to AL and used our forecasting technique to predict the performance of AL. In each step, we sampled 200 data points for \textsc{sst} and 50 data points for \textsc{subj} from the pool of unlabeled data, simulating the annotation process.  We re-trained the models in each AL step and evaluated them on the test set. \Cref{fig:case-study} demonstrates the usefulness of performance forecasting, which provides a possibility to decide on the trade-off between the additional annotation cost and the expected increase in performance.

\section{Related Work}

As the popularity of machine learning and deep learning grows, so does the need for annotated data. Since high-quality data is imperative for high-quality machine learning models, data annotation has become a lucrative industry, and new tools are constantly emerging. There are commercial tools such as Prodigy,\footnote{\url{https://prodi.gy/}} V7,\footnote{\url{https://www.v7labs.com/}} and Hasty.\footnote{\url{https://hasty.ai/}} However, these tools hide their full functionality behind a paywall.
In contrast to the mentioned commercial system, several open-source annotation tools have appeared recently, such as Label Sleuth \cite{shnarch-etal-2022-label}, Label Studio,\footnote{\url{https://labelstud.io/}}, INCEpTION \cite{klie-etal-2018-inception}, MATILDA \cite{cucurnia-etal-2021-matilda}, and Paladin \cite{nghiem-etal-2021-paladin}.

\textbf{Label Sleuth} is an elegant annotation system designed to make NLP accessible for non-experts. The system enables AL selection for labeling. However, it only supports simple binary classification with a single annotator per project.

\textbf{Label Studio} is offered as an open-source system and a paid enterprise version. While the paid version supports active learning, the free version is limited to random selection. The system supports multiple annotators but with minimal functionalities in managing the annotations.

\textbf{INCEpTION} is a highly configurable tool that supports AL and multi-annotator setups. However, the system is hard to use, as it requires external libraries to integrate a model for AL purposes.

\textbf{MATILDA} is a platform for dialogue annotation in a multi-annotator setup with support for multiple languages.

\textbf{Paladin} integrates active learning and supports multi-label classification.

To the best of our knowledge, \alanno{} is the only AL annotation tool that explicitly addresses practical challenges in AL. \alanno{} also differs from the above-mentioned systems in implementing practical solutions for managing multi-annotator annotation projects.

\section{Conclusion}
\alanno{} is an open-source annotation system for natural language processing tasks powered by active learning. The system addresses the critical practical challenges of active learning in real-world annotation projects that have previously been overlooked. \alanno{} enables non-experts in active learning to conduct effective annotation campaigns by supporting solutions for unbiased evaluation, stopping criterion for active learning, and class balancing. Additionally, the system facilitates annotation management in a multi-annotator setup, emphasizing label quality through agreement monitoring, agreement-aware label aggregation, and a novel method for the balanced assignment of unlabeled instances to annotators.
\section*{Acknowledgements}

We thank the reviewers for their comments and interesting suggestions for future improvements. We also thank the people from TakeLab (Text Analysis and Knowledge Engineering Lab) who helped us in developing the system over the years.

\bibliography{references/anthology, references/custom}
\bibliographystyle{references/acl_natbib}

\end{document}